\documentclass{article}

\usepackage{arxiv}

\usepackage[utf8]{inputenc} 
\usepackage[T1]{fontenc}    
\usepackage{hyperref}       
\usepackage{url}            
\usepackage{booktabs}       
\usepackage{amsfonts}       
\usepackage{nicefrac}       
\usepackage{microtype}      
\usepackage{lipsum}		
\usepackage{graphicx}
\usepackage{natbib}
\usepackage{doi}

\usepackage{listings}
\usepackage{array}
\usepackage{amsmath,amssymb,amsfonts}

\usepackage{algorithm}
\usepackage{algorithmicx}
\usepackage{algpseudocode}

\newenvironment{conditions*}
  {\par\vspace{\abovedisplayskip}\noindent
   \tabularx{\columnwidth}{>{$}l<{$} @{${}={}$} >{\raggedright\arraybackslash}X}}
  {\endtabularx\par\vspace{\belowdisplayskip}}

\usepackage{tabularx}

\title{HPix: Generating Vector Maps from Satellite Images}

\date{} 					

\author{Aditya~Taparia \\
	Department of Computer Science and Engineering\\
	Indian Institute of Information Technology\\
	Kottayam, Kerala, India \\
	\texttt{aditya2019@iiitkottayam.ac.in} \\
	\And
	Keshab~Nath \\
	Department of Computer Science and Engineering\\
	Indian Institute of Information Technology\\
	Kottayam, Kerala, India \\
	\texttt{keshabnath@iiitkottayam.ac.in} \\
}

\renewcommand{\shorttitle}{HPix: Generating Vector Maps from Satellite Images}

\hypersetup{
pdftitle={HPix: Generating Vector Maps from Satellite Images},
pdfauthor={Aditya~Taparia},
}

\begin{document}
\maketitle

\begin{abstract}
Vector maps find widespread utility across diverse domains due to their capacity to not only store but also represent discrete data boundaries such as building footprints, disaster impact analysis, digitization, urban planning, location points, transport links, and more. Although extensive research exists on identifying building footprints and road types from satellite imagery, the generation of vector maps from such imagery remains an area with limited exploration. Furthermore, conventional map generation techniques rely on labor-intensive manual feature extraction or rule-based approaches, which impose inherent limitations. To surmount these limitations, we propose a novel method called \textbf{HPix}, which utilizes modified Generative Adversarial Networks (GANs) to generate vector tile map from satellite images. HPix incorporates two hierarchical frameworks: one operating at the global level and the other at the local level, resulting in a comprehensive model. Through empirical evaluations, our proposed approach showcases its effectiveness in producing highly accurate and visually captivating vector tile maps derived from satellite images. We further extend our study's application to include mapping of road intersections and building footprints cluster based on their area. \textbf{GitHub: } \url{https://github.com/aditya-taparia/Satellite-Image-to-Vector-Map}.
\end{abstract}


\section{Introduction}\label{sec:intro}
Vector maps are an avant-garde representation of geographical data that transcend traditional mapping approaches. They possess unparalleled versatility, enabling the storage and depiction of discrete data boundaries, such as intricate building footprints, precise disaster impact analysis, meticulous urban planning, crucial location points, interconnected transport links, and beyond. By seamlessly integrating comprehensive information, vector maps transcend the limitations of conventional cartographic methods, empowering diverse domains with enhanced spatial understanding and decision-making capabilities.

The innovative potential of vector maps lies in their ability to encapsulate complex spatial features with utmost accuracy and fidelity. Through the intelligent manipulation of geometric primitives, including points, lines, and polygons, vector maps unlock a realm of possibilities for precise data modeling and analysis. Sophisticated algorithms and techniques facilitate the seamless conversion of raw geographical data into visually captivating and dynamically interactive vector representations, enabling efficient exploration, manipulation, and dissemination of spatial information. One of the key advantages of vector maps is their adaptability across various domains and applications. They serve as powerful tools for urban planners, aiding in the visualization of proposed developments, zoning analysis, and infrastructure management. Emergency responders leverage vector maps to swiftly assess disaster impacts, identify affected areas, and strategize rescue and recovery efforts. Transportation planners utilize vector maps to optimize routes, analyze traffic patterns, and facilitate the efficient movement of people and goods. Moreover, vector maps find extensive utility in fields such as environmental monitoring, agriculture, logistics, and public health, catalyzing data-driven decision-making and fostering innovation.

Traditional approaches to generating vector maps from satellite images have long relied on labor-intensive and time-consuming processes, often involving manual feature extraction or rule-based methodologies. These conventional techniques have inherent limitations that hinder their ability to deliver accurate and efficient results. However, recent advancements in technology and pioneering research have paved the way for newer approaches that revolutionize vector map generation. The state-of-the-art models struggle to cope with the complexity and scale of modern satellite imagery. Extracting features such as building footprints and road networks manually is a daunting task that is prone to human error and subjectivity. Additionally, rule-based approaches often require extensive prior knowledge and domain expertise, limiting their flexibility and adaptability to diverse datasets. To address these challenges, innovative newer approaches have emerged, leveraging cutting-edge technologies such as deep learning and computer vision. Convolutional Neural Networks (CNNs)\cite{iino2018cnn}, \cite{hormese2016automated}, \cite{sahu2019vector} have proven highly effective in identifying and segmenting various spatial features, including buildings, roads, and other objects of interest, from satellite images. By leveraging CNNs' capacity to learn hierarchical representations, these approaches enable automated feature extraction, reducing the reliance on labor-intensive manual efforts. Additionally, the integration of Generative Adversarial Networks (GANs)\cite{ganguli2019geogan,mansourifar2022gan} has revolutionized vector map generation, enabling the synthesis of highly accurate and visually appealing representations by mimicking the complex patterns and structures present in satellite imagery.

In this paper, we have proposed a method called HPix, short for HierarchicalPix, which utilizes generative networks to generate vector maps. By incorporating hierarchical feature learning, HPix effectively captures the intricate details of satellite imagery and translates them into accurate vector representations.

The main contribution of our paper is as follows

\begin{enumerate}
    \item We proposed an novel architecture that comprises of two GAN frameworks, one at the global level and the other at the local level. We performed an extensive assessment of \texttt{HPix} and compared with other contemporary methods.
    \item We demonstrated the versatility of \texttt{HPix} by extending its application to include mapping of road intersections and building-area clusters, showing its potential for various other use cases.
\end{enumerate}

\section{Related Work}\label{sec:related_work}
Over the past decade a lot of research has been done in different fields of satellite image processing and vector map generation. In this section, we have summarized these researches into two subsections. The first subsection discusses recent work in extracting features like road networks and building footprints from satellite images. The other subsection discusses the recent development in generating vector maps from satellite images using generative models.

\subsection{Feature Extraction from Satellite Image}
There has already been a substantial amount of work done in the field of extracting useful features from satellite images, such as road networks and building footprints. These features have been employed in various applications, including urban planning and disaster response. The idea of using artificial neural networks to extract information from satellite images was proposed as early as 2007 by the authors of \cite{bib14}. Since then, significant advancements have been made in this domain. A survey conducted by \cite{bib15} identified that in the last decade, numerous methods based on convolutional neural networks (CNN), fully convolutional networks (FCN), and U-Net architectures have been developed to extract valuable information from satellite images.

In \cite{bib13}, the authors utilized the D-LinkNet architecture to compare the effects of different loss functions on the accuracy of models for extracting road networks from satellite images. This model was chosen for its efficiency in extracting high-level information due to its encoder-decoder structure, residual blocks, and skip connections. Additionally, in research \cite{bib16}, the authors proposed a U-Net architecture with ResNet50 and compared it with other state-of-the-art models, including U-Net and Deeplabv3. They discovered that the U-Net architecture with ResNet50 provided a higher understanding of building structures and yielded better results than many existing models. While these features offer valuable information about the terrain from satellite images, obtaining a complete layout of the terrain remains a significant challenge.

\subsection{Image-to-Image Translation}
Building on the limitation of getting more comprehensive terrain information, another field of active research is in image-to-image translation. This include the use of machine learning and deep learning based algorithms, with deep learning algorithms outperforming machine learning algorithms. The deep learning based approach mainly includes Variational Auto-encoders (VAEs) and GANs \cite{bib5}. Although VAEs provided more stable training compared to GANs, several unsolved practical and theoretical challenges in VAEs boosted the use of GANs \cite{bib5}. Since GANs were first introduced in 2014 by Ian Goodfellow~\cite{bib6}, significant progress has been made in developing and enhancing various types of GANs for different image-to-image tasks.

One of the derivatives of GANs named Conditional GANs (cGANs), introduced in~\cite{bib7}, helps in controlling the output of the GANs by introducing additional information along with a random vector as input. This method was further explored by authors of~\cite{bib2}, where they introduced a conditional GAN model named Pix2Pix for various image to image translation tasks. They used an architecture inspired from U-Net for the generator and introduced a convolutional Patch-GAN classifier for the discriminator. Their research showed promising results for using cGANs for image to image translation. In 2017, another approach was introduced by the same authors for unpaired image to image translation by using  cycle consistent adversarial networks, popularly known as Cycle GANs~\cite{bib8}. Through this approach, they proposed an unpaired approach to teach cGANs to perform image to image translation and tried to solve the issue of lack of paired training data. Authors of~\cite{bib3} later worked on a similar approach as Pix2Pix with few modifications for generating vector maps from satellite images. Similarly other researches, \cite{bib9, bib10}, were also proposed for this problem statement which gave better performance than traditional methods like Cycle GAN and Pix2Pix for this problem.

\section{Methodology}\label{sec:methodology}
\begin{figure}
\centering
\includegraphics[width=0.95\linewidth]{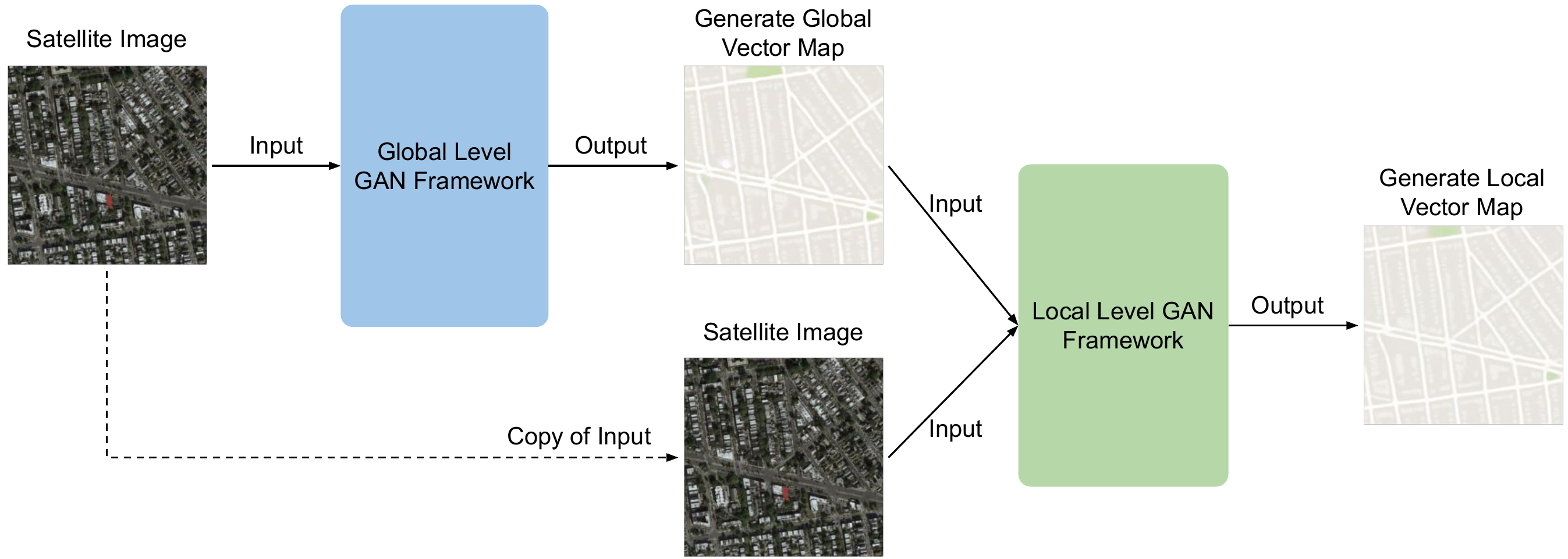}
\caption{HPix Architecture}\label{fig1}
\end{figure}

In this paper, we propose a novel architecture for translating satellite images to vector maps termed HPix (short for HierarchicalPix). This architecture comprises two GAN frameworks, one at global level and other at local level, together forming a hierarchical model, as shown in figure~\ref{fig1}. The GAN network at global level generates a coarse representation of the vector map from the input satellite image, capturing the overall layout and structure of the map. Then the GAN network at local level takes the coarse representation and the satellite image input to generate a refined version of the vector map, capturing fine-grained details and features. The local level generator also helps in reducing the artifact formation in the generated image.

The global GAN architecture comprises two components, generator and discriminator. The generator at global level comprises a complex network of encoder and decoder inspired from Unet++ architecture~\cite{bib4}, while for discriminator we used the traditional PatchGAN network, introduced in~\cite{bib2}, with slight modification in its CNN Block. The local GAN architecture also comprises two components, a generator and a discriminator. While the local discriminator is identical as global discriminator, for local generator we are using modified Pix2Pix architecture~\cite{bib2} which takes our original image along with global generated image as input to give final generated image. More detail about generator and discriminator are explained in the following subsections.

\subsection{Global Generator}
Authors of the paper \cite{bib2} explained how the use of skip connection, inspiring from Unet~\cite{bib17}, improved the output of their generator model and with that intuition we worked on improving the connection network of the generator. This generator design was inspired from Unet++ architecture~\cite{bib4}. In this architecture apart from our standard encoder-decoder network we have introduced transition blocks which take encoded data from lower level and decode them and combines that information with information from other blocks at the same level and encodes it again before passing that information further. We have also applied deep supervision to further stabilize the output of the model.

\begin{figure}
\centering
\includegraphics[width=0.95\linewidth]{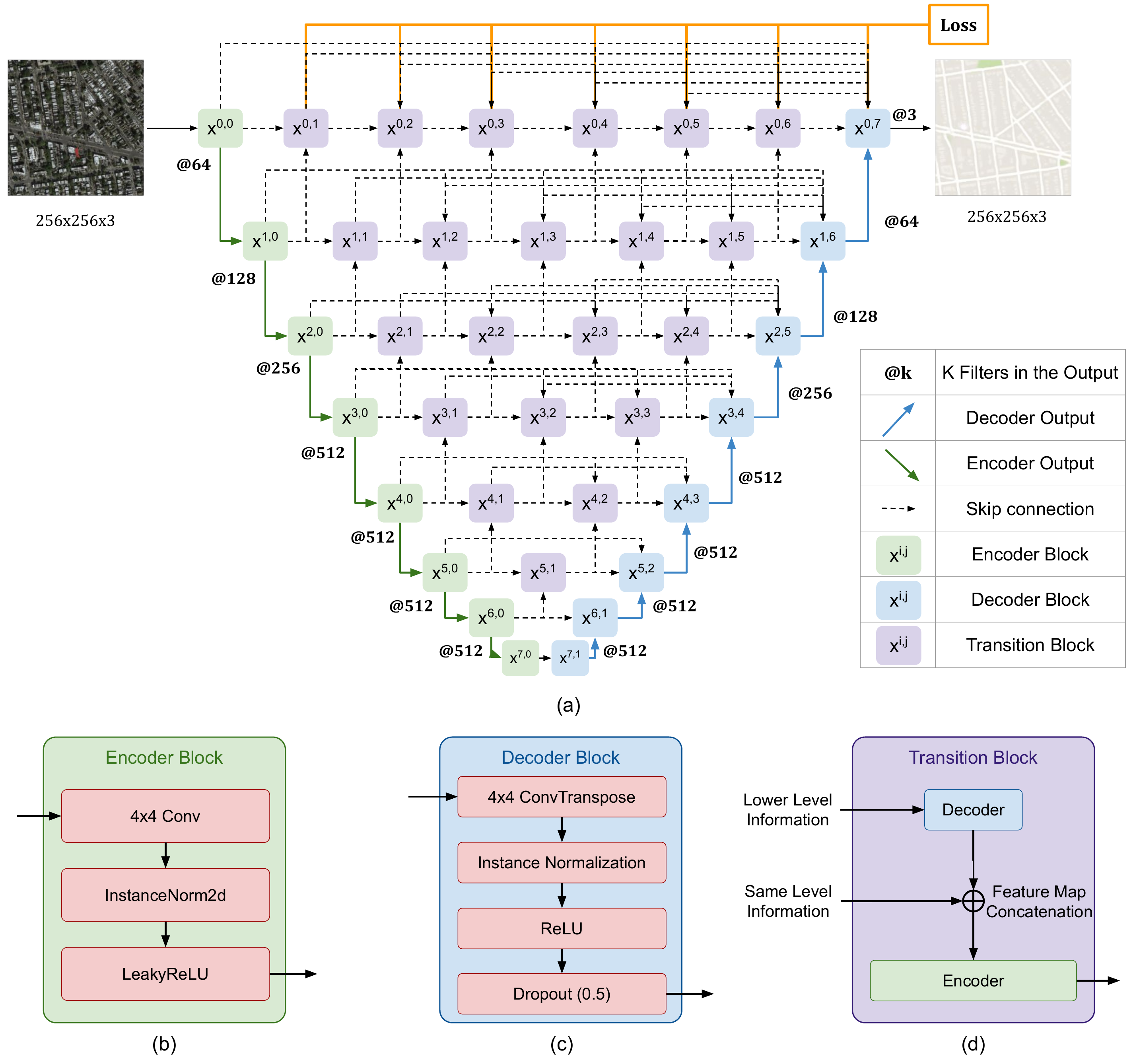}
\caption{Architecture design of (a) global generator with nested skip connection network and its components: (b) encoder block, (c) decoder block, and (d) transition block.}\label{fig2}
\end{figure}

Figure~\ref{fig2}a describes the architecture of the global generator used in HPix with the number of feature channels in the output of each encoder and decoder node. The number of features in the transition block is the same as the encoder block on that level. Figure~\ref{fig2}b, \ref{fig2}c and \ref{fig2}d  describes the architecture of encoder, decoder and transition blocks used in the network respectively. More details about the architecture is available in Appendix~\ref{app:architect}.

\subsection{Local Generator}
\begin{figure}
\centering
\includegraphics[width=0.95\linewidth]{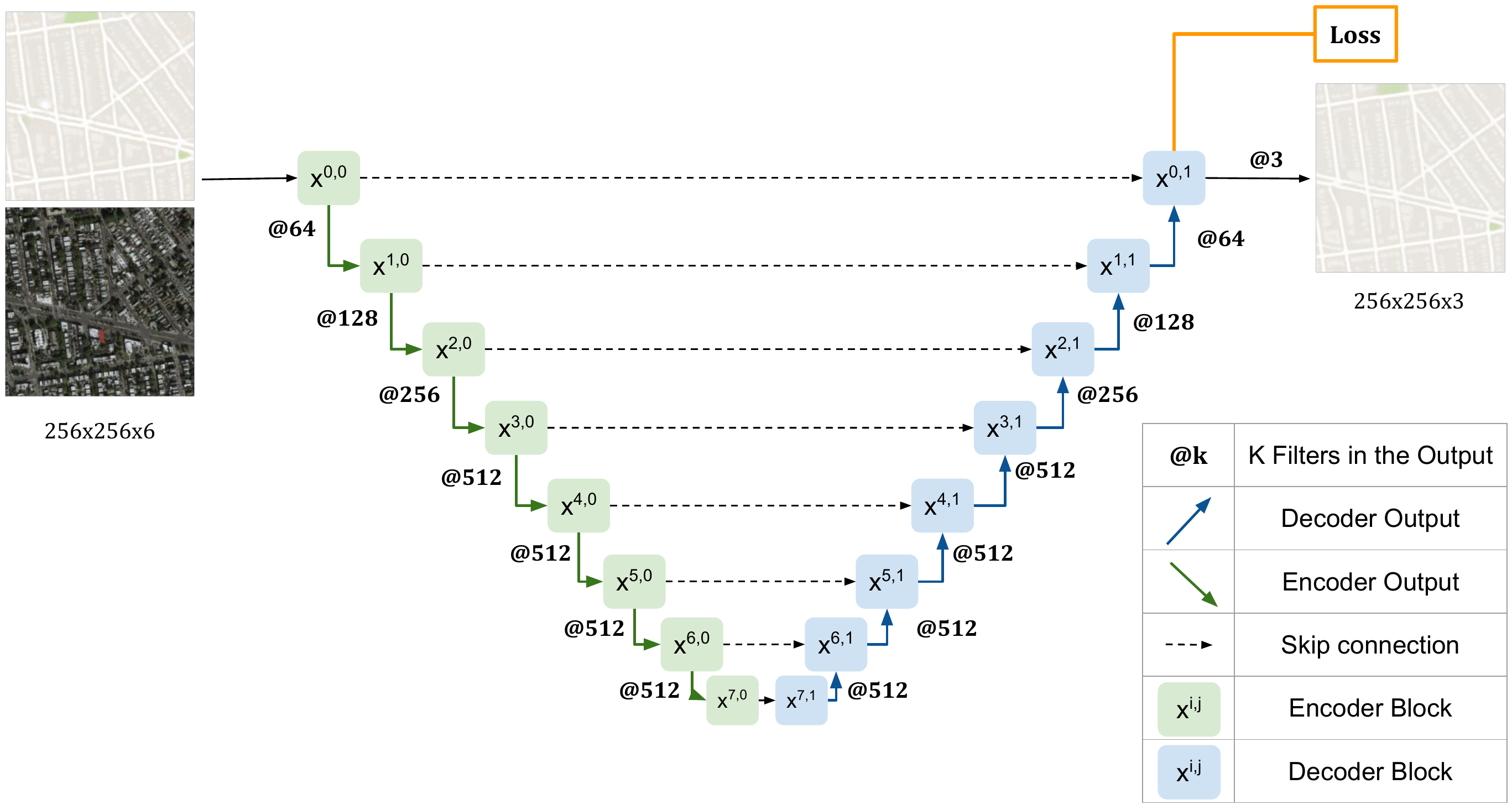}
\caption{Architecture of local generator}\label{fig3}
\end{figure}

The local generator of HierarchicalPix follows a modified architecture of Pix2Pix \cite{bib2}. It takes two inputs, a generated output of the global generator and our original input (satellite image). We identified that the use of a local generator helps in repatching some of the artifacts formed by the global generator thus improving the final output quality. Figure~\ref{fig3} describes the architecture of the local generator used in HPix with the information regarding number of feature channels in the output of each encoder and decoder node. The encoder and decoder blocks are similar in architecture to the one used in the global generator, as described in figure \ref{fig2}b and \ref{fig2}c. More details about the architecture is available in Appendix~\ref{app:architect}.

\subsection{Global and Local Discriminator}
\begin{figure}
\centering
\includegraphics[width=0.95\linewidth]{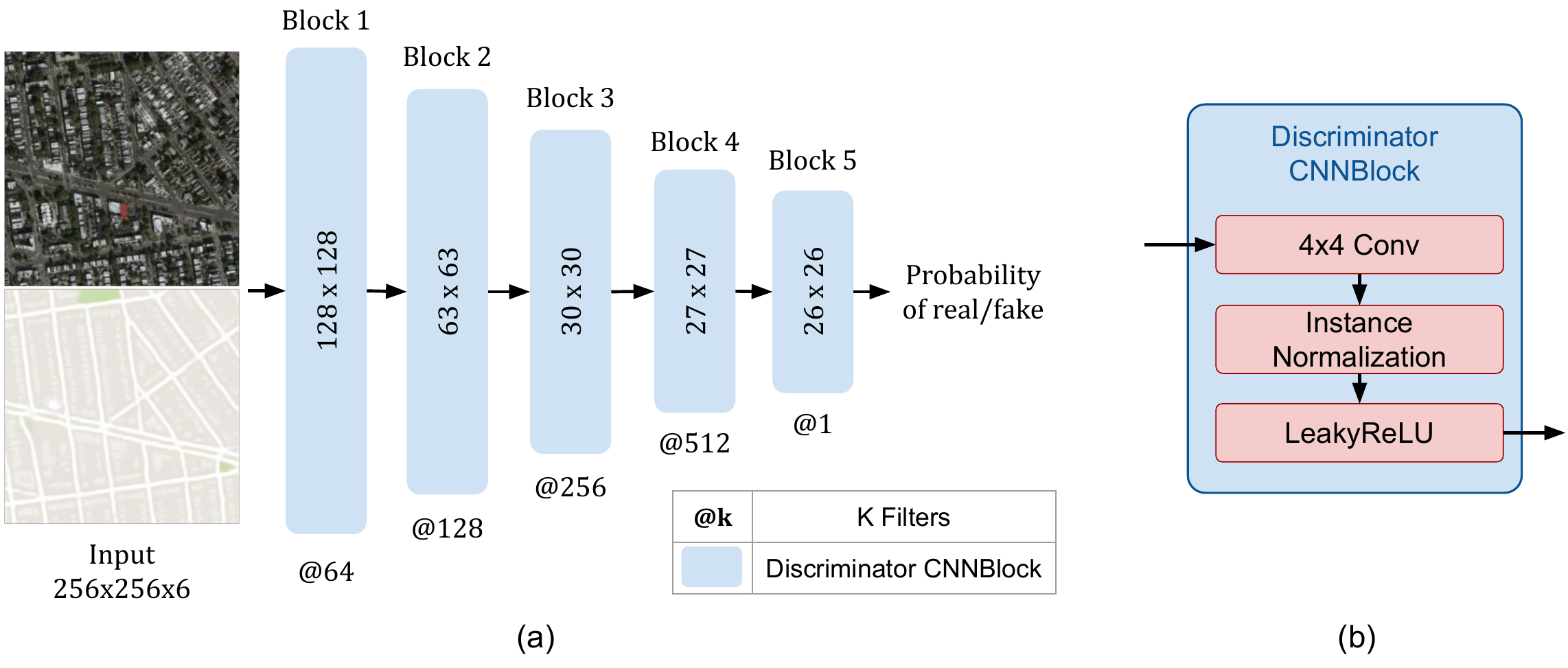}
\caption{Architecture of (a) PatchGAN discriminator and (b) its CNN block}\label{fig4}
\end{figure}

For the discriminator network, we used a 26x26 PatchGAN described in \cite{bib2}. Both the discriminator networks, global and local, are identical to each other and are used to identify real and fake images for global and local generator respectively. Figure~\ref{fig4}a shows the architecture design of the discriminator used in HPix and figure~\ref{fig4}b highlights the structure of the CNN blocks used in discriminator. More details on the architecture choices can be found in Appendix~\ref{app:architect}.

\subsection{Objective Function}
In this approach, both our GAN models are conditional GANs and the objective  of a conditional GAN can be expressed as:
\begin{equation} \label{eq1}
    \mathcal{L}_{cGAN}(G, D) = \mathbb{E}_{x, y}[\log{D(x, y)}] +  \mathbb{E}_{x,z}[1 - \log{D(x, G(x, z))}]
\end{equation}
where G tries to minimize this objective against an adversarial D and D tries to maximize this objective against an adversarial G. So, for GAN at global level the objective function can be formulated as:
\begin{equation} \label{eq2}
    \mathcal{L}_{global}(G, D_{G}) = \mathbb{E}_{x, y}[\log{D_{G}(x, y)}] +  \mathbb{E}_{x,z}[1 - \log{D_{G}(x, G(x, z))}]
\end{equation}
where G is the generator at global level, $D_{G}$ is the discriminator at global level, x is the input image, y is the ground truth or target image and z is the random noise vector. And for GAN at local level the objective function can be formulated as:
\begin{equation} \label{eq3}
    \mathcal{L}_{local}(H, D_{H}) = \mathbb{E}_{x, y}[\log{D_{H}(x, y)}] + \mathbb{E}_{x,z}[1 - \log{D_{H}(x, H(x, G(x, z), z))}]
\end{equation}
where H is the generator at local level, G is the generator at global level, $D_{H}$ is the discriminator at local level, x is the input image, y is the ground truth or target image and z is the random noise vector. Furthermore, mixing the generator objective with a traditional loss like L1 distance helps generator to not just fool the discriminator but also bring the generated output near to ground truth and generate less blurry output as highlighted in results of \cite{bib2}.
\begin{equation} \label{eq4}
    \mathcal{L}_{L1}(G) = \mathbb{E}_{x,y,z}[ \|y - G(x, z)\|]
\end{equation}
\begin{equation} \label{eq5}
    \mathcal{L}_{L1}(H) = \mathbb{E}_{x,y,z}[ \|y - H(x, G(x, z), z)\|]
\end{equation}
Our final objective functions are:
\begin{equation} \label{eq6}
    \mathcal{L}_{global}^*(G, D_{G}) = \arg\min_{G}\max_{D_{G}} \mathcal{L}_{global}(G, D_{G}) + \lambda\mathcal{L}_{L1}(G)
\end{equation}
\begin{equation} \label{eq7}
    \mathcal{L}_{local}^*(H, D_{H}) = \arg\min_{H}\max_{D_{H}} \mathcal{L}_{local}(H, D_{H}) + \lambda\mathcal{L}_{L1}(H)
\end{equation}

\section{Experimental Setup and Analysis}\label{sec:experiments}

\subsection{Dataset Acquisition and Pre-processing}
\begin{figure}
\centering
\includegraphics[width=0.95\linewidth]{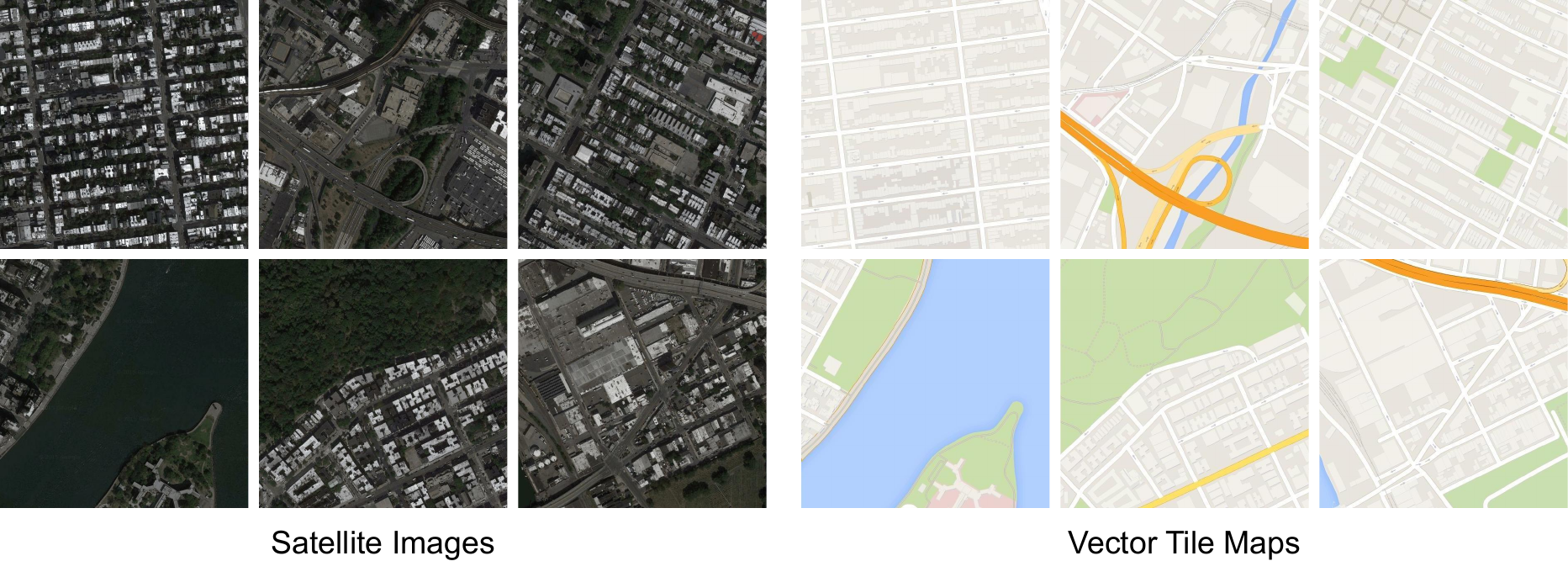}
\caption{Maps dataset sample showcasing some satellite images and their corresponding vector tile maps.}\label{fig5}
\end{figure}

We started with training the proposed network on publicly available maps dataset by \cite{bib2} and has also been used by authors of \cite{bib8, bib9, bib10} for training and testing their approaches. This dataset was collected from Google Maps and contains 1096 paired satellite and vector tile map images for training and 1098 paired satellite and vector tile map images for testing. Figure~\ref{fig5} shows some sample from this dataset.

For training and testing we resize the satellite and vector map image from 600x600 to 256x256 image. For training we also applied random jittering by first resizing the image to 286x286, then random cropping back to 256x256 sized image followed by horizontal flipping with a 50\% probability. We have also normalized both satellite and vector map images before training and testing.

\subsection{Analysis}
To compare the effectiveness of our approach with other methods, we compared them on pixel level accuracy, PSNR score and SSIM score. For calculating pixel level accuracy we considered the error factor of 5 because colors may seem similar but may vary slightly at pixel level and this strategy have been adopted in \cite{bib9, bib10, bib11} to efficiently measure this accuracy. More details on system configuration, training conditions and hyperparameters can be found in Appendix~\ref{app:exp_setup}.

\begin{figure}
\centering
\includegraphics[width=0.95\linewidth]{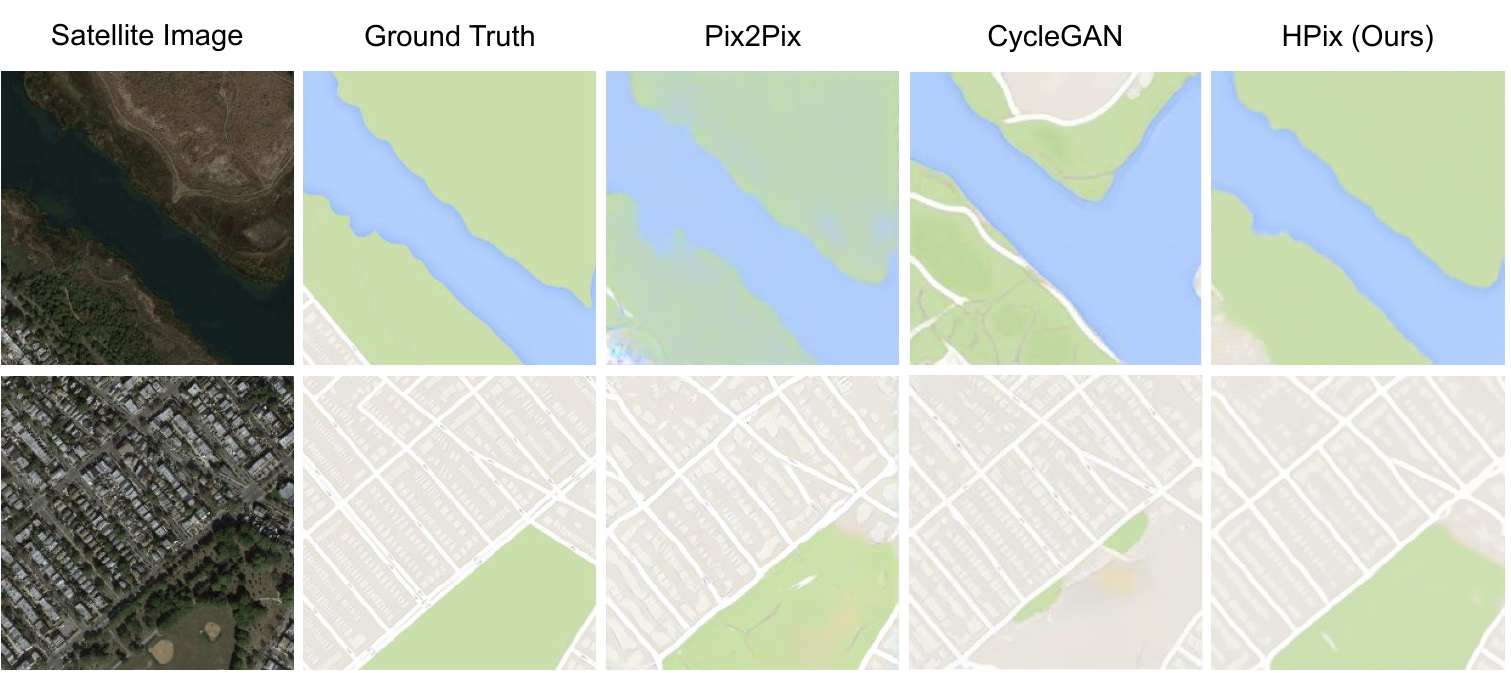}
\caption{Visual comparison of generated vector maps from satellite image by different methods on maps dataset}\label{fig6}
\end{figure}

\begin{figure}
\centering
\includegraphics[width=0.75\linewidth]{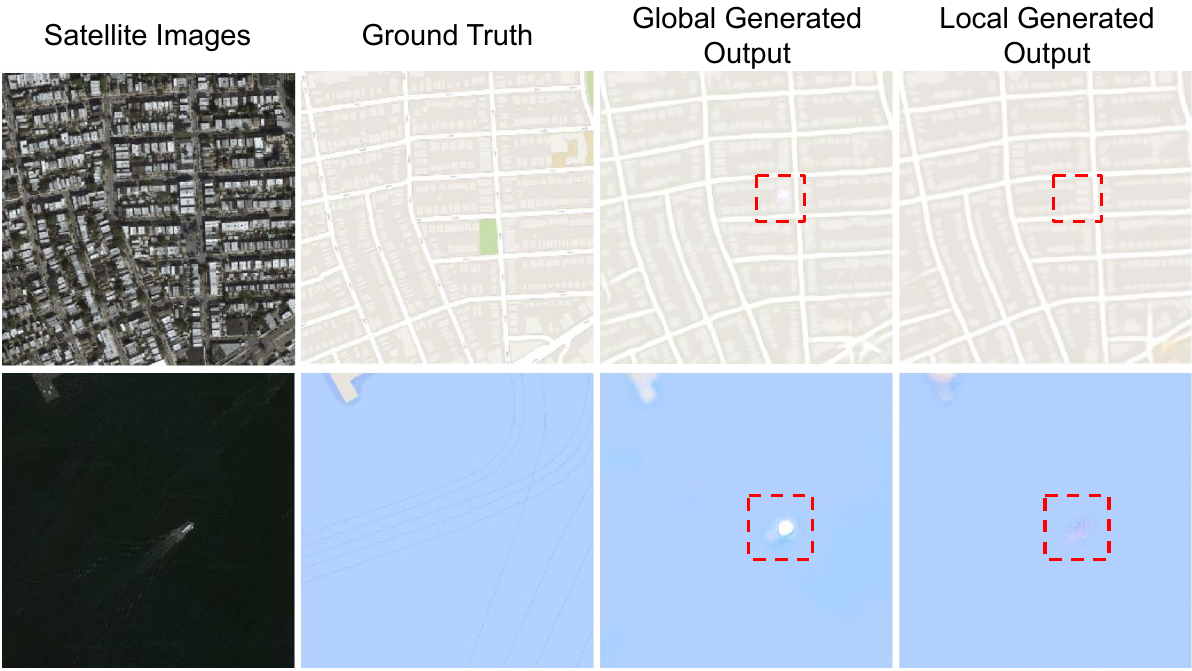}
\caption{Visualization of image generated by the global and local generator and how local generator helps in patching some of the artifacts generated by global generator.}\label{fig7}
\end{figure}

Table~\ref{tab1} shows the comaprision of our approach with other methodologies. In our comparision, we considered Pix2Pix \cite{bib2} and CycleGAN \cite{bib8} as our baseline. Furthermore, we also included algorithms from recent research including CscGAN \cite{bib9} and MapGen-GAN \cite{bib10}. From the experimental analysis we concluded that our approach performs better on most of the metrics when compared with other models and the PSNR score of our approach is almost comparable to CscGAN (current best). Figure~\ref{fig6} shows the visual comparison between output generated by PixPix, CycleGAN and our approach as compared to ground truth. Figure~\ref{fig7} displays how using the local generator helped in patching up the artifacts generated by the global generator and improving the overall output quality.

\begin{table}
    \caption{Comparative analysis of different models with best result highlighted in bold.}
    \centering
    \begin{tabular}{@{}llll@{}}
        \toprule
        Model & Pixel level accuracy & SSIM score & PSNR score \\
        \midrule
        Pix2Pix & 42.09\% & 0.64 & 25.29 \\
        CycleGAN & 36.47\% & 0.63 & 24.05 \\
        MapGen-GAN & 38.54\% & 0.64 & 24.64 \\
        CscGAN & 46.86\% & 0.73 & \textbf{27.20} \\
        HPix (Ours) & \textbf{61.04\%} & \textbf{0.75} & 26.98 \\
        \bottomrule
    \end{tabular}
    \label{tab1}
\end{table}

\section{Use cases}\label{sec:usecase}
\begin{figure}
\centering
\includegraphics[width=0.95\linewidth]{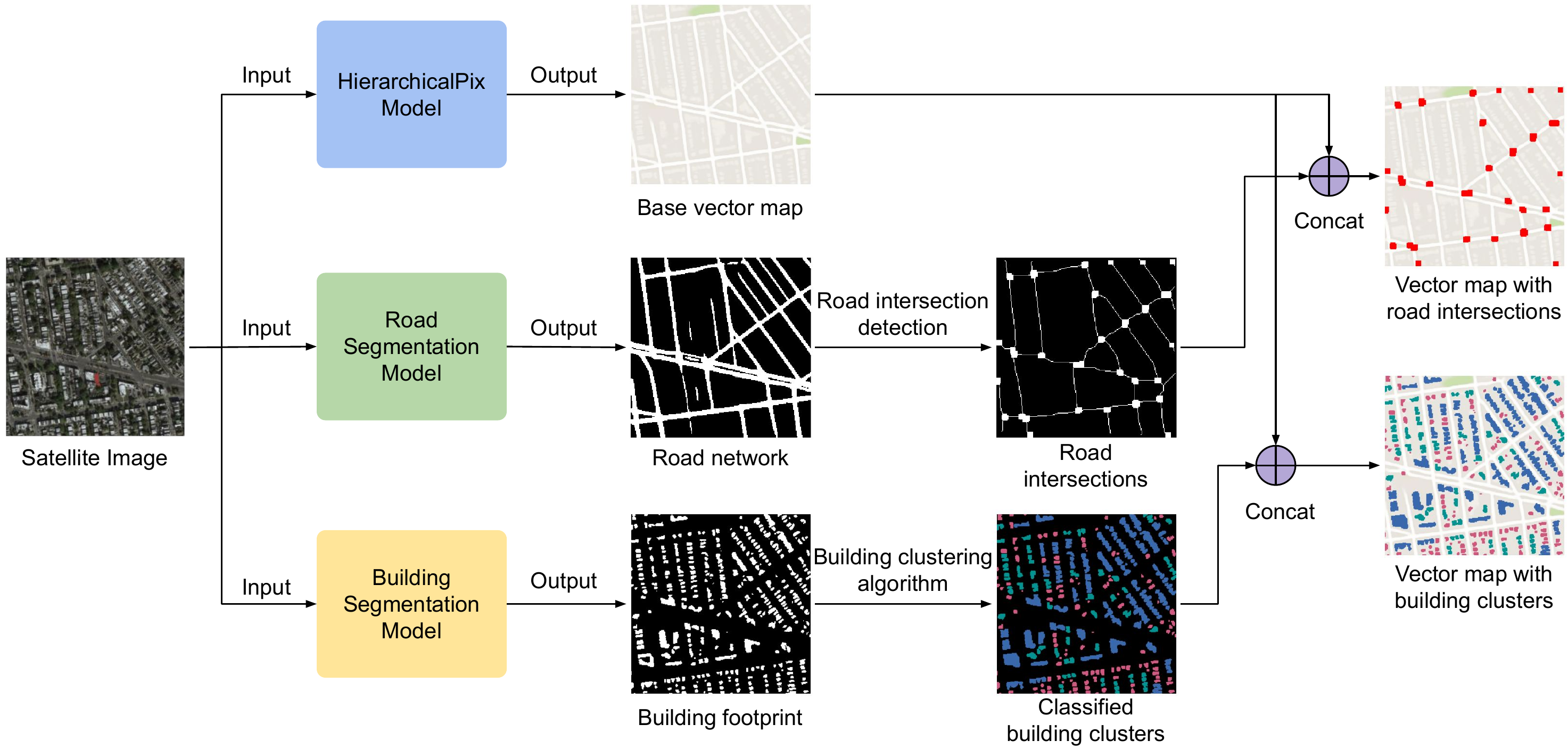}
\caption{Mapping other applications with generated output from HPix to create interactive vector maps.}\label{fig8}
\end{figure}

\begin{figure}
\centering
\includegraphics[width=0.95\linewidth]{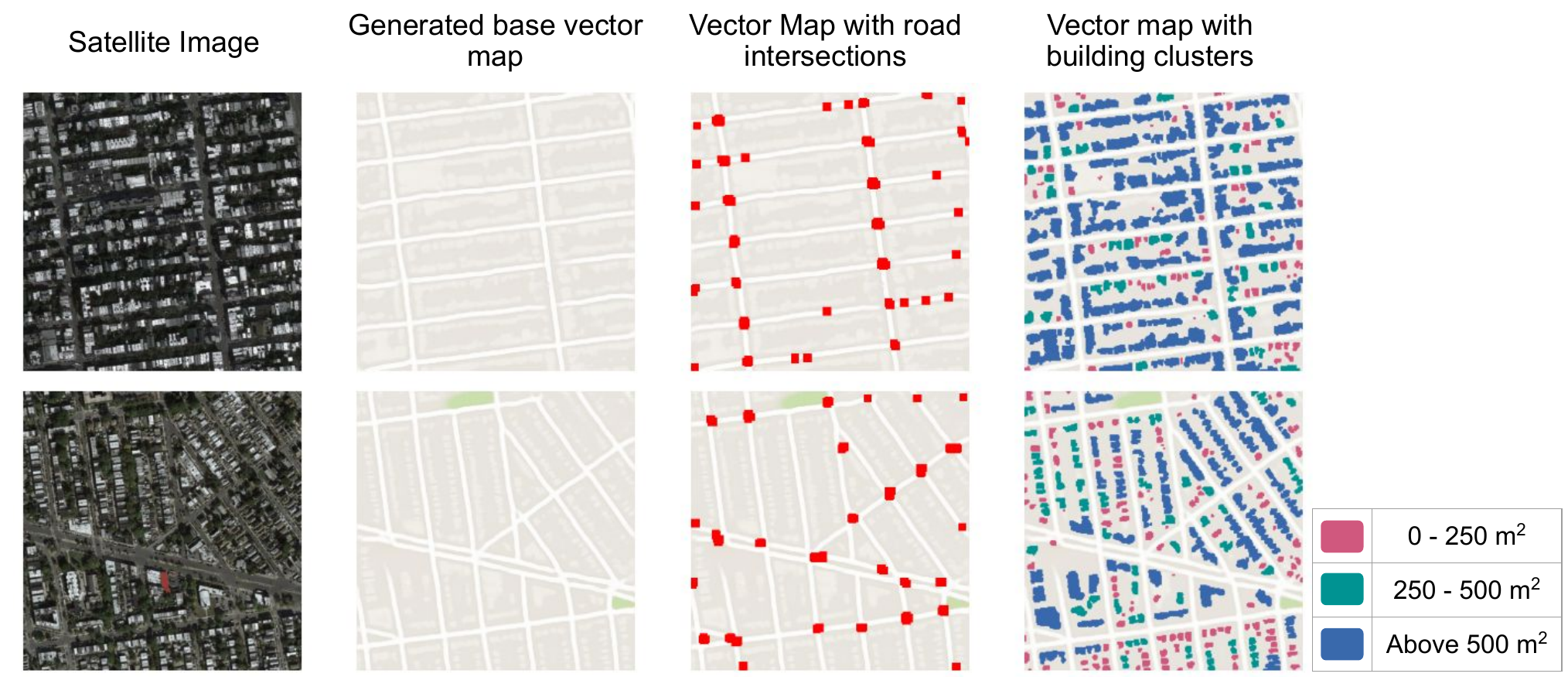}
\caption{Some samples of generated vector map with marked road intersection and classified building clusters based on area.}\label{fig9}
\end{figure}

We further extended the study to include mapping of road intersections and building clusters, from the satellite image to generated vector map. We accomplish this by first segmenting road networks and building footprints from the satellite image. This is then followed by running road intersection detection and building cluster identification algorithm, as described in Appendix~\ref{app:usecase}. After identification of road intersections and building clusters based on area, we combine the generated vector map with these information to generate interactive vector maps. The satellite images have resolution of 1 meter per pixel and buildings are highlighted with area between 0 and 250 square meters as red, area between 250 and 500 square meters as green and any building with area above 500 square meters as blue. Figure~\ref{fig8} highlights the overall flow of the approach and figure~\ref{fig9} shows some sample generated from this approach. More details on the models used and algorithm followed in  use cases can be found in Appendix~\ref{app:usecase}.

\section{Conclusion}\label{sec:conclusion}
In this paper, we have proposed a novel method for generating vector tile map from satellite image termed HPix. This architecture comprises of two generators, global and local, for identifying complex features in the input image and map it with ground truth. We have also found that using local generator helps in reducing the number of artifacts in the generated output, thus improving the overall generated output quality. The experimental results show that our model produces better translation results than other state-of-the-art approaches on maps dataset. We later use this generated vector map to create interactive vector map by marking road intersections and separating building clusters based on size highlighting the use cases of our method. The architectural design of our approach promotes its use as a general-purpose solutions like edges-to-photo, BW-to-color, or labels-to-street scene.

\bibliographystyle{unsrtnat}
\bibliography{references}  

\begin{thebibliography}{22}
\providecommand{\natexlab}[1]{#1}
\providecommand{\url}[1]{\texttt{#1}}
\expandafter\ifx\csname urlstyle\endcsname\relax
  \providecommand{\doi}[1]{doi: #1}\else
  \providecommand{\doi}{doi: \begingroup \urlstyle{rm}\Url}\fi

\bibitem[Iino et~al.(2018)Iino, Ito, Doi, Imaizumi, and Hikosaka]{iino2018cnn}
Shota Iino, Riho Ito, Kento Doi, Tomoyuki Imaizumi, and Shuhei Hikosaka.
\newblock Cnn-based generation of high-accuracy urban distribution maps utilising sar satellite imagery for short-term change monitoring.
\newblock \emph{International journal of image and data fusion}, 9\penalty0 (4):\penalty0 302--318, 2018.

\bibitem[Hormese and Saravanan(2016)]{hormese2016automated}
Jose Hormese and C~Saravanan.
\newblock Automated road extraction from high resolution satellite images.
\newblock \emph{Procedia Technology}, 24:\penalty0 1460--1467, 2016.

\bibitem[Sahu and Ohri(2019)]{sahu2019vector}
M~Sahu and A~Ohri.
\newblock Vector map generation from aerial imagery using deep learning.
\newblock \emph{ISPRS Annals of the Photogrammetry, Remote Sensing and Spatial Information Sciences}, 4:\penalty0 157--162, 2019.

\bibitem[Ganguli et~al.(2019)Ganguli, Garzon, and Glaser]{ganguli2019geogan}
Swetava Ganguli, Pedro Garzon, and Noa Glaser.
\newblock Geogan: A conditional gan with reconstruction and style loss to generate standard layer of maps from satellite images.
\newblock \emph{arXiv preprint arXiv:1902.05611}, 2019.

\bibitem[Mansourifar et~al.(2022)Mansourifar, Moskowitz, Klingensmith, Mintas, and Simske]{mansourifar2022gan}
Hadi Mansourifar, Alex Moskowitz, Ben Klingensmith, Dino Mintas, and Steven~J Simske.
\newblock Gan-based satellite imaging: A survey on techniques and applications.
\newblock \emph{IEEE Access}, 2022.

\bibitem[Mokhtarzade and Zoej(2007)]{bib14}
Mehdi Mokhtarzade and MJ~Valadan Zoej.
\newblock Road detection from high-resolution satellite images using artificial neural networks.
\newblock \emph{International journal of applied earth observation and geoinformation}, 9\penalty0 (1):\penalty0 32--40, 2007.

\bibitem[Chen et~al.(2022)Chen, Deng, Luo, Li, Junior, Gon{\c{c}}alves, Nurunnabi, Li, Wang, and Li]{bib15}
Ziyi Chen, Liai Deng, Yuhua Luo, Dilong Li, Jos{\'e}~Marcato Junior, Wesley~Nunes Gon{\c{c}}alves, Abdul Awal~Md Nurunnabi, Jonathan Li, Cheng Wang, and Deren Li.
\newblock Road extraction in remote sensing data: A survey.
\newblock \emph{International Journal of Applied Earth Observation and Geoinformation}, 112:\penalty0 102833, 2022.

\bibitem[Xu et~al.(2023)Xu, He, Zhang, Ma, and Li]{bib13}
Hongzhang Xu, Hongjie He, Ying Zhang, Lingfei Ma, and Jonathan Li.
\newblock A comparative study of loss functions for road segmentation in remotely sensed road datasets.
\newblock \emph{International Journal of Applied Earth Observation and Geoinformation}, 116:\penalty0 103159, 2023.

\bibitem[Alsabhan and Alotaiby(2022)]{bib16}
Waleed Alsabhan and Turky Alotaiby.
\newblock Automatic building extraction on satellite images using unet and resnet50.
\newblock \emph{Computational Intelligence and Neuroscience}, 2022, 2022.

\bibitem[Pang et~al.(2021)Pang, Lin, Qin, and Chen]{bib5}
Yingxue Pang, Jianxin Lin, Tao Qin, and Zhibo Chen.
\newblock Image-to-image translation: Methods and applications.
\newblock \emph{IEEE Transactions on Multimedia}, 24:\penalty0 3859--3881, 2021.

\bibitem[Goodfellow et~al.(2020)Goodfellow, Pouget-Abadie, Mirza, Xu, Warde-Farley, Ozair, Courville, and Bengio]{bib6}
Ian Goodfellow, Jean Pouget-Abadie, Mehdi Mirza, Bing Xu, David Warde-Farley, Sherjil Ozair, Aaron Courville, and Yoshua Bengio.
\newblock Generative adversarial networks.
\newblock \emph{Communications of the ACM}, 63\penalty0 (11):\penalty0 139--144, 2020.

\bibitem[Mirza and Osindero(2014)]{bib7}
Mehdi Mirza and Simon Osindero.
\newblock Conditional generative adversarial nets, 2014.
\newblock Preprint at \url{https://arxiv.org/abs/1411.1784}.

\bibitem[Isola et~al.(2017)Isola, Zhu, Zhou, and Efros]{bib2}
Phillip Isola, Jun-Yan Zhu, Tinghui Zhou, and Alexei~A Efros.
\newblock Image-to-image translation with conditional adversarial networks.
\newblock In \emph{Proceedings of the IEEE conference on computer vision and pattern recognition}, pages 1125--1134, 2017.

\bibitem[Zhu et~al.(2017)Zhu, Park, Isola, and Efros]{bib8}
Jun-Yan Zhu, Taesung Park, Phillip Isola, and Alexei~A Efros.
\newblock Unpaired image-to-image translation using cycle-consistent adversarial networks.
\newblock In \emph{Proceedings of the IEEE international conference on computer vision}, pages 2223--2232, 2017.

\bibitem[Ingale et~al.(2021)Ingale, Singh, and Patwal]{bib3}
Vaishali Ingale, Rishabh Singh, and Pragati Patwal.
\newblock Image to image translation: Generating maps from satellite images, 2021.
\newblock Preprint at \url{https://arxiv.org/abs/2105.09253}.

\bibitem[Liu et~al.(2021)Liu, Wang, Fang, Zhou, Sun, Zheng, and Chen]{bib9}
Yuanyuan Liu, Wenbin Wang, Fang Fang, Lin Zhou, Chenxing Sun, Ying Zheng, and Zhanlong Chen.
\newblock Cscgan: Conditional scale-consistent generation network for multi-level remote sensing image to map translation.
\newblock \emph{Remote Sensing}, 13\penalty0 (10):\penalty0 1936, 2021.

\bibitem[Song et~al.(2021)Song, Li, Chen, and Wu]{bib10}
Jieqiong Song, Jun Li, Hao Chen, and Jiangjiang Wu.
\newblock Mapgen-gan: a fast translator for remote sensing image to map via unsupervised adversarial learning.
\newblock \emph{IEEE Journal of Selected Topics in Applied Earth Observations and Remote Sensing}, 14:\penalty0 2341--2357, 2021.

\bibitem[Zhou et~al.(2018)Zhou, Rahman~Siddiquee, Tajbakhsh, and Liang]{bib4}
Zongwei Zhou, Md~Mahfuzur Rahman~Siddiquee, Nima Tajbakhsh, and Jianming Liang.
\newblock Unet++: A nested u-net architecture for medical image segmentation.
\newblock In \emph{Deep Learning in Medical Image Analysis and Multimodal Learning for Clinical Decision Support: 4th International Workshop, DLMIA 2018, and 8th International Workshop, ML-CDS 2018, Held in Conjunction with MICCAI 2018, Granada, Spain, September 20, 2018, Proceedings 4}, pages 3--11. Springer, 2018.

\bibitem[Ronneberger et~al.(2015)Ronneberger, Fischer, and Brox]{bib17}
Olaf Ronneberger, Philipp Fischer, and Thomas Brox.
\newblock U-net: Convolutional networks for biomedical image segmentation.
\newblock In \emph{Medical Image Computing and Computer-Assisted Intervention--MICCAI 2015: 18th International Conference, Munich, Germany, October 5-9, 2015, Proceedings, Part III 18}, pages 234--241. Springer, 2015.

\bibitem[Fu et~al.(2019)Fu, Gong, Wang, Batmanghelich, Zhang, and Tao]{bib11}
Huan Fu, Mingming Gong, Chaohui Wang, Kayhan Batmanghelich, Kun Zhang, and Dacheng Tao.
\newblock Geometry-consistent generative adversarial networks for one-sided unsupervised domain mapping.
\newblock In \emph{Proceedings of the IEEE/CVF Conference on Computer Vision and Pattern Recognition}, pages 2427--2436, 2019.

\bibitem[Demir et~al.(2018)Demir, Koperski, Lindenbaum, Pang, Huang, Basu, Hughes, Tuia, and Raskar]{bib19}
Ilke Demir, Krzysztof Koperski, David Lindenbaum, Guan Pang, Jing Huang, Saikat Basu, Forest Hughes, Devis Tuia, and Ramesh Raskar.
\newblock Deepglobe 2018: A challenge to parse the earth through satellite images.
\newblock In \emph{Proceedings of the IEEE Conference on Computer Vision and Pattern Recognition Workshops}, pages 172--181, 2018.

\bibitem[Mnih(2013)]{bib12}
Volodymyr Mnih.
\newblock \emph{Machine learning for aerial image labeling}.
\newblock University of Toronto (Canada), 2013.

\end{thebibliography}

\newpage
\appendix
\onecolumn
\section*{Appendix}
In this appendix, we show the network architectures, experimental setup, algorithms used and additional figures.

\section{Network architectures}\label{app:architect}

\subsection{Global generator network}
The encoder block comprises a Conv-InstanceNorm-LeakyReLU layer. The first encoder block ($x_{0,0}$) doesn’t apply InstanceNorm to its convoluted output and the bottleneck encoder block ($x_{7,0}$) doesn’t apply InstanceNorm and LeakyReLU to its convoluted output. Our decoder block comprises a ConvTranspose-InstanceNorm-ReLu layer which is followed by a dropout layer with 50\% probability. The final decoder block ($x_{0,7}$) doesn’t apply InstanceNorm and LeakyReLU to its convoluted output, instead just applies Tanh. We have used skip connections to pass the feature map information between encoder, decoder and transition blocks. While applying deep supervision, output from blocks $x_{0,1}$, $x_{0,2}$, $x_{0,3}$, $x_{0,4}$, $x_{0,5}$ and $x_{0,6}$ are first passed through a convolution layer followed by Tanh and are then considered for calculating loss.

In this network, we have used an instance norm layer rather than a batch norm layer and applied reflection padding to reduce the appearance of artifacts in the generated output \cite{bib8}.

\subsection{Local generator network}
The first encoder block ($x_{0,0}$) doesn’t apply InstanceNorm to its convoluted output and the bottleneck encoder block ($x_{7,0}$) doesn’t apply InstanceNorm and LeakyReLU to its convoluted output. The final decoder block ($x_{0,1}$) doesn’t apply InstanceNorm and LeakyReLU to its convoluted output, instead just applies Tanh. We have used skip connections to pass the feature map information between encoder, decoder.

\subsection{Discriminator network}
The discriminator blocks consist of Convolution-InstanceNorm-LeakyReLU layers. The first layer (block 1) doesn’t have an InstanceNorm layer and we used LeakyReLU with a slope of 0.2. The last block, block 5, doesn’t have an InstanceNorm and LeakyReLU layer and applies convolution with stride 1. The output from block 5 is a single channel output.

\section{Experimental setups}\label{app:exp_setup}
\subsection{Computing resources}
The training of models were performed on the Kaggle platform by using community-available two Nvidia Tesla T4 GPUs with 13 GB RAM and 2 CPU cores. It took around 16 hours to train the model and around 10 minutes to validate the trained model. The code was written using Pytorch library in python.

\subsection{Hyperparameters and training conditions}
While training HPix, we have trained both generators simultaneously so that they could learn and generalize the problem together. We used Adam optimizer for both the generators and discriminators with a learning rate of 0.0002 and beta1 and beta2 as 0.5 and 0.999. We trained the model on objective function defined in methodology for 200 epochs.

\section{Use cases}\label{app:usecase}
\subsection{Road network and intersection}
\begin{figure}
\centering
\includegraphics[width=0.65\linewidth]{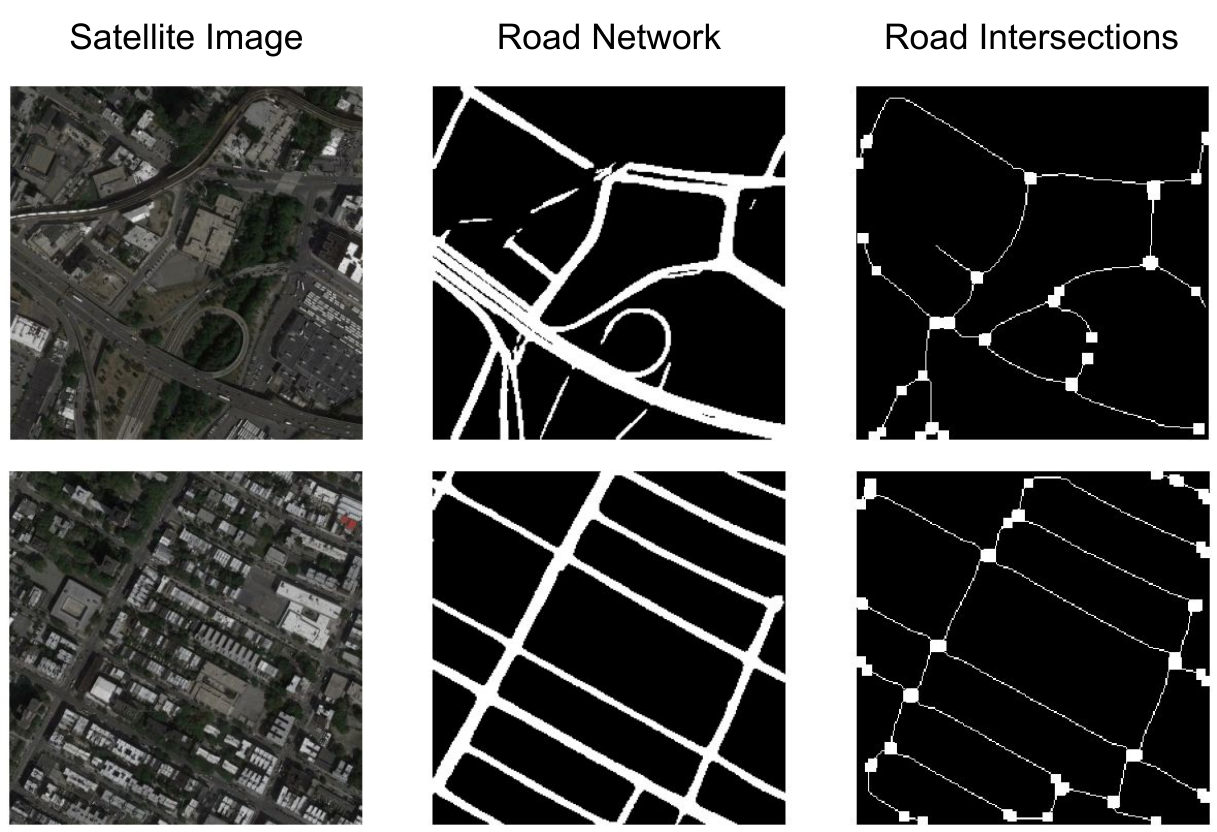}
\caption{Road network and intersection mapping using pre-trained model on maps dataset.}\label{fig10}
\end{figure}

For extracting road network from the satellite image, we used a pre-trained DLinkNet model trained on DeepGlobe Road Extraction Dataset \cite{bib19}. The generated binary segmented map of the road network is then used with algorithm~\ref{alg1} for identifying road intersections. Figure~\ref{fig10} show samples of identified road network and their intersections.

\begin{algorithm}
\caption{Road intersection detection}\label{alg1}
\begin{algorithmic}[1]
    \Procedure{RoadIntersection}{$Map$} \Comment{Map: 3D array of pixels}
        \State $GrayscaleMap \gets ConverttoGrayscale(Map)$
        
        \State $BinaryMap \gets ConverttoBinary(GrayscaleMap)$ 

        \State $GaussianMap \gets GaussianBlur(BinaryMap, kernel=(31 \times 31))$ 

        \State $DilatedMap \gets GaussianMap$

        \For{$i \gets 0$ to $5$}                    
            \State $DilatedMap \gets Dilate(DilatedMap, kernel=(3 \times 3))$ 
        \EndFor

        \State $ThresholdedMap \gets Threshold(DilatedMap, threshold=25)$

        \State $ErodedMap \gets ThresholdedMap$

        \For{$i \gets 0$ to $5$}                    
            \State $ErodedMap \gets Erode(ErodedMap, kernel=(3 \times 3))$ 
        \EndFor

        \State $SkeletonMap \gets Skeletonize(ErodedMap)$ 

        \State $CornerMap \gets FindCorners(SkeletonMap)$ 

        \State $IntersectionMap \gets DrawCorners(Map, CornerMap)$ 

        \State \Return $IntersectionMap$ \Comment{IntersectionMap: 2D array of pixels}
    \EndProcedure
\end{algorithmic}
\end{algorithm}

\subsection{Building footprint and clusters}
\begin{figure}
\centering
\includegraphics[width=0.85\linewidth]{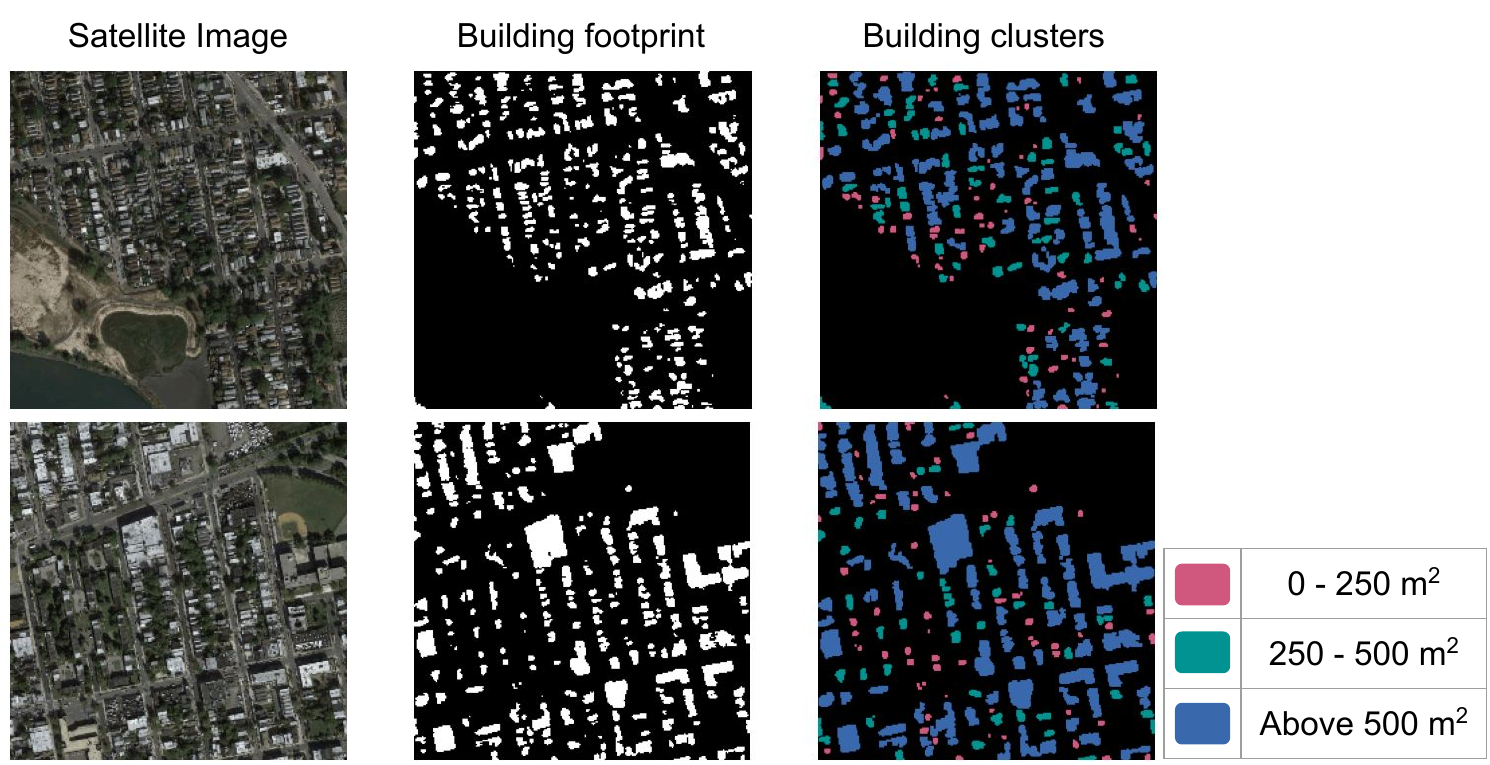}
\caption{Building footprint detection using trained Unet++ model and cluster classification based on area.}\label{fig11}
\end{figure}

We first extracted the building footprints from the satellite image by training a Unet++ \cite{bib4} model on Massachusetts building dataset from \cite{bib12}. Then we used algorithm~\ref{alg2} for classifying buildings in that segmented map. Figure~\ref{fig11} show samples of building footprint and their clusters based on area.

\begin{algorithm}
\caption{Building classification}\label{alg2}
\begin{algorithmic}[1] 
    \Procedure{BuildingClassifier}{$Map, Resolution, Thresholds, ColorLabels$} \Comment{Map: 2D array of pixels, Resolution: meters per pixel, Thresholds: small, medium, and large building sizes, ColorLabels: small, medium, and large building colors}

        \State $Counters \gets FindContours(Map)$

        \State $Areas \gets CalculateAreas(Counters, Resolution)$

        \For{$i \gets 0$ to $len(Areas)$}                    
            \If {$Areas[i] < SmallThreshold$}
                \State $Labels \gets DrawContours(Map, Counters[i], $1$)$
            \ElsIf {$Areas[i] < MediumThreshold$}
                \State $Labels \gets DrawContours(Map, Counters[i], $2$)$
            \Else
                \State $Labels \gets DrawContours(Map, Counters[i], $3$)$
            \EndIf
        \EndFor

        \State $ClassifiedMap \gets RecolorMap(Labels, ColorLabels)$
        
        \State \Return $ClassifiedMap$ \Comment{ClassifiedMap: 2D array of pixels}
    \EndProcedure
\end{algorithmic}
\end{algorithm}

\end{document}